\pdfoutput=1

\documentclass[11pt]{article}
\usepackage{array}
\newcolumntype{P}[1]{>{\raggedright\arraybackslash}p{#1}}
\usepackage[dvipsnames]{xcolor}
\usepackage{amsmath}

\usepackage[final]{acl}

\usepackage{times}
\usepackage{latexsym}

\usepackage[T1]{fontenc}

\usepackage[utf8]{inputenc}

\usepackage{microtype}

\usepackage{inconsolata}

\usepackage{graphicx}

\title{AutoSpec: An Agentic Framework for Automatically Drafting Patent Specification}

\author{Ryan Shea \\
  Columbia University, NY \\
  \texttt{rs4235@columbia.edu} \\\And
  Zhou Yu \\
  Columbia University, NY \\
  \texttt{zy2461@columbia.edu} \\}

\begin{document}
\maketitle
\begin{abstract}
Patents play a critical role in driving technological innovation by granting inventors exclusive rights to their inventions. However the process of drafting a patent application is often expensive and time-consuming, making it a prime candidate for automation. Despite recent advancements in language models, several challenges hinder the development of robust automated patent drafting systems. First, the information within a patent application is highly confidential, which often prevents the use of closed-source LLMs for automating this task. Second, the process of drafting a patent application is difficult for even the most advanced language models due to their long context, technical writing style, and specialized domain knowledge. To address these challenges, we introduce AutoSpec, a secure, agentic framework for \textbf{Auto}matically drafting patent \textbf{Spec}ification. Our approach decomposes the drafting process into a sequence of manageable subtasks, each solvable by smaller, open-source language models enhanced with custom tools tailored for drafting patent specification. To assess our system, we design a novel evaluation protocol in collaboration with experienced patent attorneys. Our automatic and expert evaluations show that AutoSpec outperforms existing baselines on a patent drafting task.
\end{abstract}


\section{Introduction}

Drafting a patent application has long been a key component of intellectual property protection. Yet, the drafting process remains a difficult and laborious task. Inventors face many hurdles to patenting their inventions, including high monetary costs and significant time commitments \cite{wang-etal-2024-patentformer}. This discourages smaller entities and individual inventors from pursuing patents, stifling innovation and competition.

LLMs offer a promising way to alleviate these issues by automating the patent drafting process. Recent work has shown that LLMs can achieve impressive performance on many complex writing tasks, including ones within the legal domain \cite{katz2023legal, ariai2025legal}. However, there remain several obstacles to the development and deployment of automatic patent drafting systems.

A major challenge in automating patent drafting is ensuring the security and confidentiality of sensitive invention details. Leakage of this information could compromise the patent’s validity or result in an outright rejection of the application. This makes on-premises deployment of patent drafting systems highly desirable, restricting the use of more powerful, proprietary LLMs for solving this task.

These challenges are compounded by the fact that patent drafting remains difficult for even the most advanced language models \cite{Jiang_2025_patent_survey}. Patent specifications often span tens of thousand of words, which is beyond what current LLMs can output in a single generation. Patent applications also integrate highly specialized domain knowledge, using a combination of legal and technical language that is difficult for LLMs to replicate \cite{wang-etal-2024-patentformer}.

\begin{figure*}[h]
    \centering
    \includegraphics[width=\textwidth]{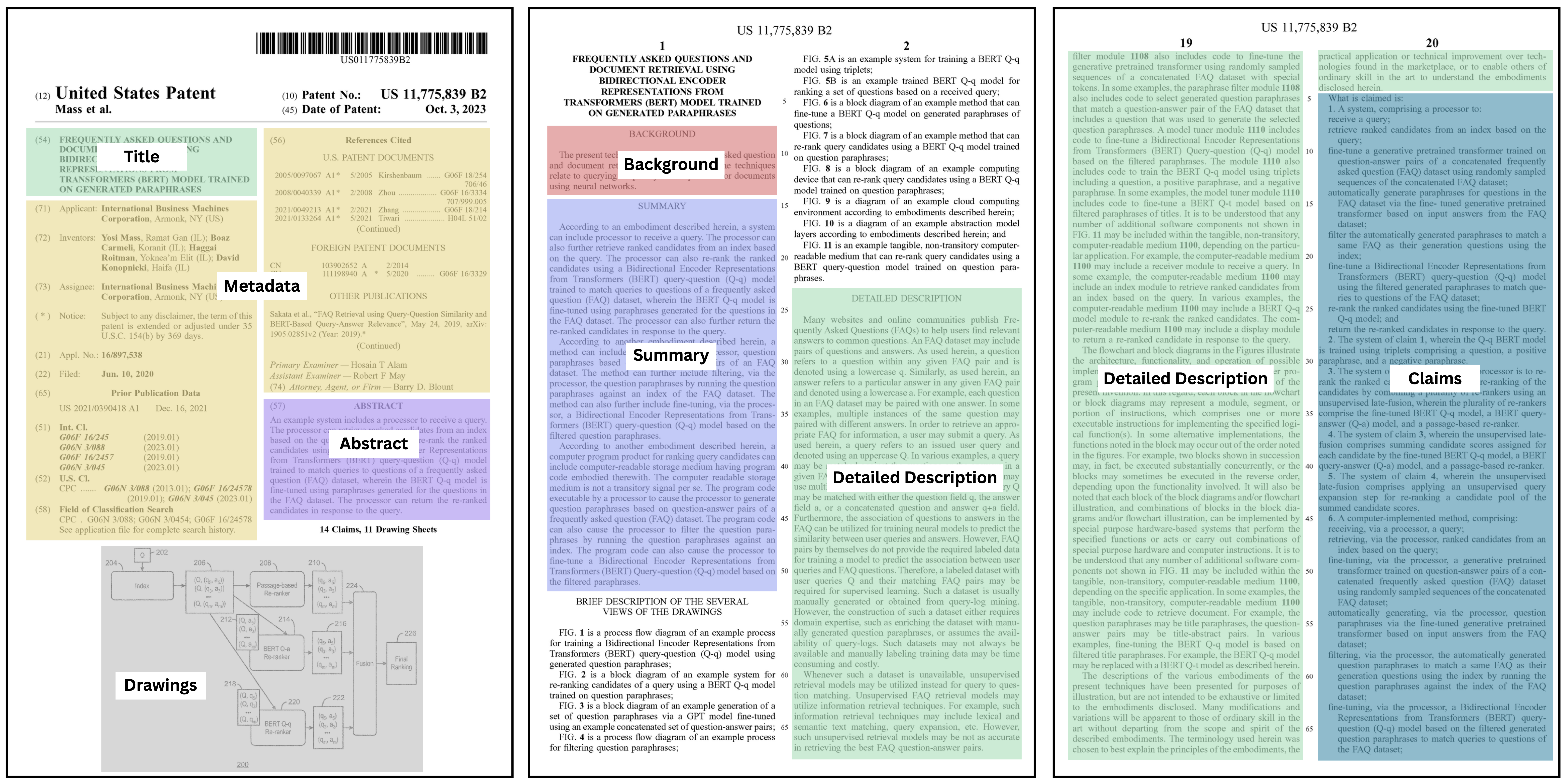}
    \caption{Three pages of a published patent application. The specification consists of the Abstract, Background, Summary, and Detailed Description.}
    \label{fig:patent_diagram}
\end{figure*}

To address these issues we propose AutoSpec, an agentic method for \textbf{Auto}matically generating patent application \textbf{Spec}ification. Given the core details of an invention, AutoSpec produces a full specification by first creating a structured outline. Our outline generation method is constructed to emulate the way that patent attorneys decompose the workflow for drafting patent specification. This outline breaks down the drafting process into manageable subtasks, each of which are solvable by smaller, open-source LLMs in combination with custom-built tools we create specifically for drafting patent specification. These custom tools are designed according to expert input and use a combination of fine-tuning, prompting, and retrieval to effectively draft patent application disclosure.

To rigorously assess our system, we introduce a novel evaluation protocol for analyzing patent specifications, developed in collaboration with expert patent attorneys. This protocol is centered on an annotation scheme designed to highlight the critical aspects of high-quality patent disclosure and to standardize the evaluation of machine-generated patent applications. Leveraging this protocol, we evaluate our approach on a patent drafting task using both automated metrics and human expert assessments. Our results show that AutoSpec outperforms existing baselines. We release our evaluation data which consists of 75 machine-generated patent disclosures annotated according to our evaluation protocol. Our contributions are summarized as follows:
\begin{itemize}
\item We introduce AutoSpec, a novel agentic framework for drafting patent specification. AutoSpec is built around open-source LLMs, ensuring drafting remains secure and reliable.
\item We design an evaluation protocol for evaluating patent disclosure, developed with expert input to capture the key elements of high-quality patent specification.
\item We evaluate our framework on a patent drafting task. Our results show that AutoSpec outperforms existing baselines according to automatic and expert evaluations.
\end{itemize}

\section{Background}
Patent applications are legal documents that define an invention. They consist of a set of claims and a specification (also referred to as the disclosure). The claims serve to define the scope of an invention concisely and unambiguously. The specification is written based on the content in the claims, and typically includes an abstract, background, summary, and detailed description. Depending on the field of the invention, the patent specification may also contain drawings and drawing descriptions which are included in the detailed description. Figure \ref{fig:patent_diagram} shows a complete example of a patent application.

Patent language is often very technical, using specialized terms, legal jargon, and sometimes new terms to describe novel concepts. Patents frequently create their own definitions for terms, which can differ significantly from how those words are used in normal language or even within the relevant technical field. These self-defined terms are typically quite artificial and are unlikely to appear in other documents. This makes it difficult for LLMs, which are typically trained on general internet text, to emulate the kind of language in patent applications \cite{Jiang_2025_patent_survey}. 

Patent specifications are very long with a length of about 13.5k tokens on average \cite{hupd_neurips}. The detailed description comprises the bulk of this length with an average of 11.9k tokens. The specification is written primarily based on claim information which is typically around 1.3k tokens. This means the disclosure must elaborate heavily on the content in the claims and incorporate relevant external concepts that are key to explaining the invention. This combination of long generation and elaboration is also difficult for LLMs which are not extensively trained on long text sequences. For example, despite it's 128k context length LLaMA 3 expends only about 5.5\% of it's computational budget training on text sequences longer than 8k tokens \cite{grattafiori2024llama3herdmodels, llama_paper}.

Patents are granted on the basis of novelty, meaning that if any information about the invention is in the public domain the patent application will be rejected. This limits the use of proprietary LLMs due to privacy concerns \cite{li-etal-2025-papillon}. Instead, on premises deployment with open-source models is desirable. However, these models tend to be less capable which further exacerbates difficulty of automatically drafting patent disclosure. 

\section{Related Work}
Prior work in the field of patent generation has typically been centered around generating shorter sections of the specification such as the abstract or summary \cite{jiang2024largelanguagemodelsgenerate, zhou2024masterslaveencodermodelimproving}. Some additional works have proposed tasks such as next claim generation or creating individual figure descriptions \cite{aubakirova2023patfiggeneratingshortlong, shukla2025patentlmmlargemultimodalmodel, lee2019patentclaimgenerationfinetuning, jiang-etal-2025-large}. Other recent directions of research include paraphrasing disclosure and simplifying/revising claims \cite{Casola_2023_patent_summarization, jiang-etal-2025-patent-revision}.

The closest related method to ours is Patentformer \cite{wang-etal-2024-patentformer}. In this work, the authors fine-tune a GPT-J and T5 \cite{raffel2023_t5} model on pairs of claims and their closest related paragraph in the specification. While this method does allow models to generate patent disclosure, it makes several simplifying assumptions that diverge from actual patent drafting practices, for instance, the notion that each paragraph in the specification maps to a single claim \cite{Jiang_2025_patent_survey}. Training on pairs of claims and their closest matching paragraph also disincentivizes the model from elaborating. Patent disclosure often reiterates the claims to some extent, which may encourage the model to simply reiterate claim content without including the external information needed for drafting the full disclosure.

Prior work on evaluating machine-generated patent disclosure has been limited. Most existing works use metrics such as perplexity or BLEU for evaluation \cite{papineni-etal-2002-bleu}. While this can be somewhat effective for short texts, these measures have been shown to struggle evaluating longer sequences \cite{hu2024perplexitylongtext}. The most comprehensive work in this area is PatentEval \cite{zuo-etal-2024-patenteval} which proposes an error typology for generating new claims based on previous ones and for generating the abstract based on the claims. To our knowledge, there is no existing protocol for evaluating full patent specifications.

\begin{figure*}[h]
    \centering
    \includegraphics[width=\textwidth]{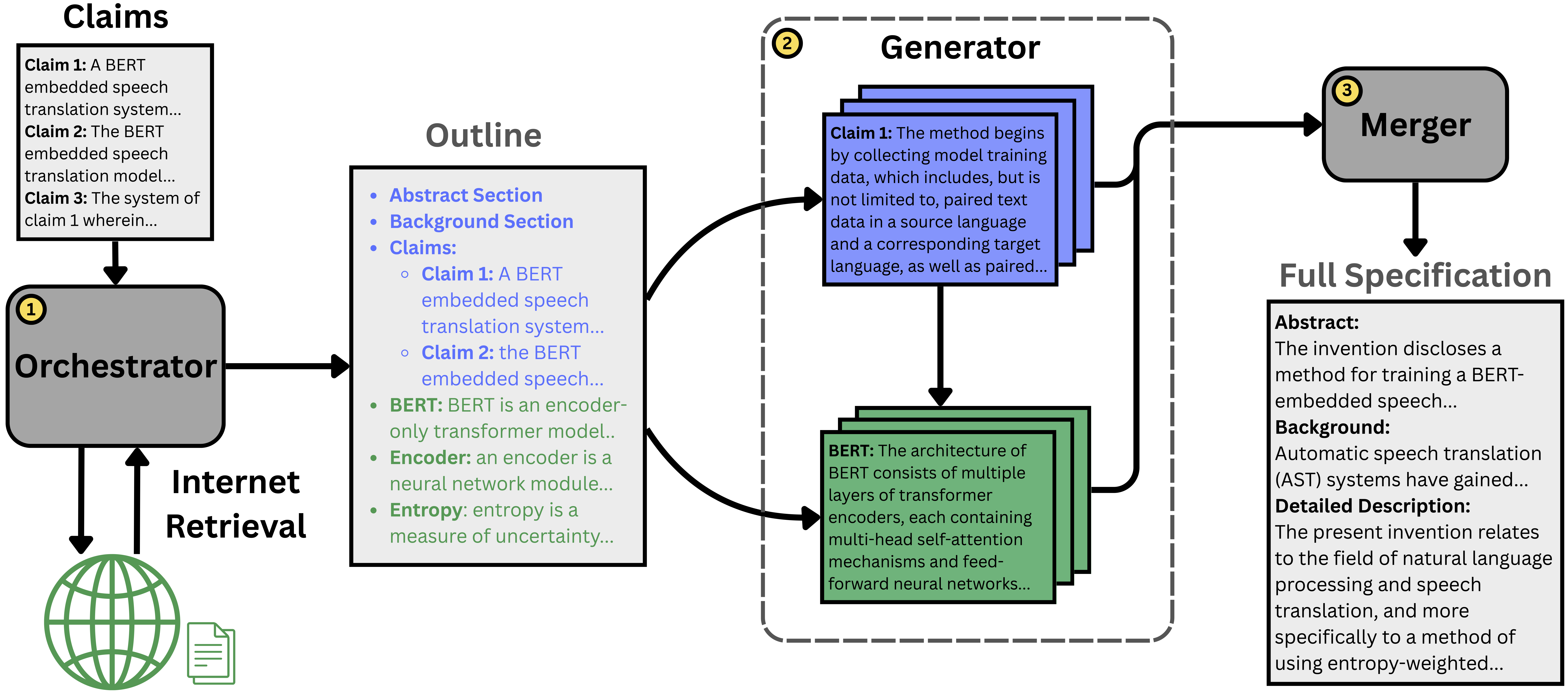}
    \caption{A diagram illustrating the workflow of AutoSpec. The orchestrator uses the claims, OCR-extracted figure text, and an internet search tool to generate an outline consisting of \textcolor{NavyBlue}{template items} and \textcolor{OliveGreen}{technical items}. The generator creates the specification for each outline item, then the merger combines them to form the full specification.}
    \label{fig:patent_associate}
\end{figure*}

\section{Method}
\label{sec:method}

In this section we outline AutoSpec, a novel agentic method for generating full patent specifications. Our method is designed to take in the claims of a patent application, along with optional OCR-extracted figure text, and generate the specification. AutoSpec consists of three main components: the \textbf{orchestrator}, \textbf{generator}, and \textbf{merger}. An overview of the AutoSpec workflow is illustrated in Figure \ref{fig:patent_associate}.

\subsection{Orchestrator}

The orchestrator is designed to process the claims and OCR-extracted figure text of a patent in order to generate an outline for the full disclosure. It starts by building a template composed of standard components found in most patent applications. These include shorter sections such as the abstract and background, as well as the claims themselves. We refer to the items added in this stage as ``template items.''

The orchestrator then expands the initial template by adding items specific to the particular patent application. This is done by prompting an open-source LLM to extract key technical concepts from the claims, along with some brief information about each concept. Depending on the length and complexity of the claims, this extraction may be performed in a single pass or across multiple iterations. Each item added to the outline in this stage is marked as requiring a retrieval step. We refer to the items added in this stage as ``technical items.''

For each technical item, the orchestrator uses an internet search api to retrieve relevant information about the concept. This internet search api can be a proprietary tool as the individual technical concepts do not contain any sensitive information that would compromise the integrity of the invention. The retrieved documents, combined with the original claims, are then passed to a language model, which generates contextually relevant content that aligns with the invention as described in the claims. This output is appended to the corresponding technical item in the outline.  This step is essential for ensuring the disclosure meaningfully expands upon the claims rather than merely restating them. We observe that open-source language models often struggle to elaborate effectively on claim content without the aid of external information.

This orchestration process reflects how patent attorneys typically approach drafting disclosures. Attorneys typically write patent applications in segments, some of which are more standardized and primarily involve restating or discussing the claims, while others require more detailed explanation and the inclusion of external information. This approach forms the basis for the two item categories in our structured outline: template items for standard content and technical items for concept-specific elaboration.

\subsection{Generator}

The generator is responsible for producing all of the text that appears in the final disclosure. It is built based on an open-source LLM that has been trained on patent specifications to better capture the language and style typical of patent applications. This domain-specific training is crucial for ensuring the generated text aligns with standard drafting conventions. The generator powers two custom tools, each designed to handle a different type of item in the outline.

The first tool is designed to generate the specification for each template item in the outline. Since these sections can be written using only the claims and figure information, the tool takes as input the claims, the relevant outline item, and a custom prompt. It then uses the generator to produce the corresponding portion of the disclosure.

The second tool is responsible for generating the disclosure sections corresponding to the technical items in the outline. It takes as input a custom prompt, the claims, the specific outline item, and the existing disclosure content produced by the first tool. Including the previously generated disclosure provides valuable context, enabling the model to produce more coherent and relevant text that better aligns with content of the specification. 

The specification for each item in the outline is generated using one of the two tools. The disclosure for all template items must be completed first, as it serves as input for generating the content of the technical items. This ordering mirrors the workflow of human drafters, who often begin with standardized sections before elaborating on specific technical details.

\subsection{Merger}

The merger is designed to take all of the output given by the generator and combine them to create the final specification. The sections corresponding to each template item are produced independently and merged simply by concatenating the subsections in order. After merging, each paragraph is sequentially numbered. An LLM is then used to integrate the disclosure for the technical items. The model is prompted to provide reasoning about where to insert the paragraph, indicate the insertion position, and generate a revised version of the paragraph to ensure a smooth transition between sections.

\section{Evaluation Protocol}
\label{sec:annotation_scheme}

To establish a consistent framework for evaluating patent disclosures, we developed a novel evaluation protocol in collaboration with experienced patent attorneys. This protocol is based on an annotation scheme that identifies key elements of high-quality patent specification. Disclosures are assessed across five categories, each rated on a scale from one to five. The category definitions and scoring guidelines are detailed below.

\noindent
\textbf{Language style} evaluates how closely the language and word choice of the specification matches the style a human would use when writing a description of an invention. The disclosure should be dry and factual, avoiding excessive promotional or advocating language. It should not directly reference the claims (for example, by saying ``as given in claim 1'') and should avoid "patent profanity," which are overly specific words like "crucial" or "critical" when describing the invention. These terms unnecessarily limit the scope of the invention and reduce its enforceability.

A score of one reflects pervasive issues, including excessive advocacy, use of patent profanity, and frequent claim references. A score of three indicates a mix of inappropriate and acceptable language, with substantial portions written in a suitable legal tone. A score of five signifies that the specification is almost entirely written in the proper style, using dry, factual language with minimal issues. Minimal advocating language is acceptable.

\noindent
\textbf{Elaboration} assesses how well the specification expands on the content of the claims. Good disclosure should not simply repeat the claim language but should explain and elaborate on the key technical concepts contained in the claims to help the reader better understand the scope of invention.

A score of one indicates that the specification simply restates the claims with little or no additional detail. A score of three suggests that some elaboration is present, but much of the disclosure closely mirrors the claim content. A score of five reflects thorough elaboration on all key technical concepts needed to adequately understand the invention.

\noindent
\textbf{Diversity score} evaluates the diversity of language and content in the disclosure. Good specification should not repeat the same content or unnecessarily extend its length by restating the same points. It should also avoid repeating long strings of text and should use some variation in language throughout.

A score of one indicates a high level of repetition, with substantial duplication of content or long strings of identical text. A score of three suggests moderate repetition, though significant portions of the specification show adequate variation. A score of five indicates that the specification has little to no unnecessary repetition, resembling the writing style of a human.

\noindent
\textbf{Factual accuracy} evaluates how accurate the content in the disclosure is. All of the content should be factual without hallucinations. The disclosure should also not contain any references to nonexistent figures, claims, other sections not present in the patent specification.

A score of one indicates frequent inaccuracies, including false or misleading statements and references to nonexistent elements. A score of three reflects occasional issues, but the majority of the content is accurate and consistent with the claims. A score of five signifies that the disclosure is entirely accurate, with no hallucinated content or invalid references.

\noindent
\textbf{Coverage of claims} evaluates whether or not there is any content missing in the disclosure. All of the claims should be addressed in the disclosure without any important information omitted.

A score of one indicates that the specification covers few, if any, of the claims, with significant omissions. A score of three means that only partial coverage is provided, approximately half of the claim content is addressed. A score of five indicates that the specification fully incorporates all information from the claims, with no important elements missing.

\begin{table*}[ht]
\centering
\begin{tabular}{P{3.5cm} P{2.5cm} P{2.9cm} P{2.05cm} P{2.cm}}
\hline
\textbf{Model} & \textbf{PatentSBERTa Similarity$\uparrow$} & \textbf{BERT for Patents Similarity$\uparrow$} & \textbf{Patent Profanity$\downarrow$} & \textbf{Diversity Difference$\downarrow$}\\
\hline
LLaMA 3.3 & 0.821 (0.078) & 0.931 (0.042) & 0.44 (0.70) & 1.23 (1.22)\\
GPT-4o (Single-Gen) & 0.834 (0.070) & 0.925 (0.046) & 0.92 (0.97)	& 1.90 (1.07) \\
GPT-4o (Multi-Gen) & \textit{0.866 (0.064)} & \textit{0.944 (0.037)} & 33.70 (19.94) & 1.47 (1.80)\\
Patentformer & 0.821 (0.083) & 0.941 (0.036) & \textbf{0.02  (0.14)} & 3.87 (3.49)\\
AutoSpec (Template) & 0.835 (0.076) & 0.943 (0.039) & \textit{0.21 (0.64)} & \textbf{0.11  (1.69)} \\
AutoSpec (Ours) & \textbf{0.879 (0.071)} & \textbf{0.950 (0.037)} & 0.28 (0.75) & \textit{1.23 (1.18)}\\
\hline
\end{tabular}
\caption{\label{tab:auto_eval}
Automatic evaluation results of AutoSpec against four baselines along with an alternative version of AutoSpec with only template items in the outline. The best scores for each category are \textbf{bold}, the second best scores are shown in \textit{italics}, standard deviations are in parenthesis.
}
\end{table*}

\section{Experiments}
We evaluate the effectiveness of our agent through both automated metrics and expert human assessments. The results demonstrate that AutoSpec outperforms multiple baseline approaches in generating patent specifications. Additionally, we perform an error analysis comparing AutoSpec with two baseline models, using feedback provided by our expert evaluators. To support further research, we release the expert evaluation dataset, which includes 75 machine-generated patent disclosures annotated according to the protocol described in Section \ref{sec:annotation_scheme}.

\subsection{Implementation}

For our implementation of AutoSpec we utilize LLaMA 3.3 70b as the base LLM. We fine-tune this model using LoRA \cite{hu2021loralowrankadaptationlarge} on data consisting of claim-specification pairs. Our data is sourced from a subset of the HUPD dataset \cite{hupd_neurips} supplemented by data scraped from Google patents. The HUPD dataset only includes patent applications up to 2018, so we collect this additional data in order to incorporate more recent patent specifications. We make this data publicly available for replication purposes.

The orchestrator, generator, and merger all use this trained model for their tasks. We find that LoRA fine-tuning allows the model to better replicate patent language while retaining it's general-purpose instruction-following capabilities. Additional details for our implementation can be found in Appendix \ref{appendix:implementation}.

\subsection{Baselines}

We evaluate our method against four baseline approaches, described below. Further implementation details for each method are in Appendix \ref{appendix:implementation}.

\textbf{LLaMA 3.3} We use the LLaMA 3.3 70b parameter base model as our first baseline. This model has no fine-tuning or any of the additional components given in Section \ref{sec:method}.

\textbf{GPT-4o (Single-Gen)} For this baseline, we prompt GPT-4o to generate the entire patent disclosure in a single pass. The input includes only the template items defined in Section \ref{sec:method}, without any technical items.

\textbf{GPT-4o (Multi-Gen)} This baseline also uses GPT-4o, but generates the disclosure section by section. For each template item, the model is prompted using the current item along with previously generated content to maintain coherence across the full draft.

\textbf{Patentformer} The final baseline is based on the Patentformer method introduced by \citet{wang-etal-2024-patentformer}. We fine-tune the LLaMA 3.3 70B model on their released dataset of claim-specification pairs. Disclosure is generated iteratively, with the model conditioned on the claims and the previously generated paragraph.

\begin{table*}
\centering
\begin{tabular}{p{3.1cm}p{2.5cm}p{2cm}p{1.9cm}p{2cm}p{1.9cm}}
\hline
\textbf{Method} & \textbf{Language Style} & \textbf{Elaboration} & \textbf{Diversity} & \textbf{Factual Acc.} & \textbf{Coverage} \\
\hline
GPT-4o (Multi-Gen) & 3.24 (0.60) & \textbf{3.68 (0.63)$^{*}$} & 3.08 (1.00)\textsuperscript{\textdagger} & 3.92 (0.57)\textsuperscript{\textdagger} & 3.84 (0.75)\textsuperscript{\textdagger} \\
Patentformer & 3.80 (1.08)\textsuperscript{\textdagger} & 2.20 (1.00) & 2.28 (1.10) & 2.84 (1.40) & 1.96 (0.98) \\
AutoSpec & \textbf{3.96 (0.68)\textsuperscript{\textdagger}} & 3.24 (0.88)\textsuperscript{\textdagger} & \textbf{3.60 (0.87)$^{*}$} & \textbf{4.04 (0.98)\textsuperscript{\textdagger}} & \textbf{4.32 (0.99)$^{*}$} \\
\hline
\end{tabular}
\caption{\label{tab:patent_eval}
Expert evaluation results of GPT-4o (Multi-Gen), Patentformer, and AutoSpec. The best scores for each category are shown in \textbf{bold}, standard deviations are in parenthesis. Statistically significant improvements (independent two-sample t-test, $p < 0.05$) over both baselines are marked with $^*$, improvements over one baseline are marked with \textsuperscript{\textdagger}.
}
\end{table*}

\subsection{Automatic Evaluation}

For our automatic evaluation we generated patent disclosures for 100 published patents selected by two patent experts in the field of biotechnology. The generated disclosures were assessed using the following metrics. Additional implementation details are provided in Appendix \ref{appendix:implementation}.

\textbf{Semantic Similarity} We assess semantic similarity using two embedding models specifically trained for use on patent text: PatentSBERTa \cite{BEKAMIRI2024_patentsberta} and BERT for Patents\footnote{\href{https://github.com/google/patents-public-data/blob/master/models/BERT\%20for\%20Patents.md}{https://github.com/google/patents-public-data}}. For each model, we compute embeddings for both the original and generated disclosures, and calculate cosine similarity to measure alignment.

\textbf{Patent Profanity} We approximate the language quality of each method by checking for the presence of patent profanity within the disclosure. This is done via keyword matching using a curated list of problematic terms and phrases provided by patent experts.

\textbf{N-gram Diversity} We use n-gram diversity from \citealp{li-etal-2016-diversity-ngrams} to estimate the linguistic variety within each disclosure. Patents naturally include some repetition, so we report the absolute difference in average n-gram diversity between the generated disclosure and the original specification to capture language diversity.

The results of our automatic evaluations are presented in Table \ref{tab:auto_eval}. We also include results for a variant of AutoSpec that only uses the template items from the outline to generate the disclosure. AutoSpec achieves the highest scores on both semantic similarity metrics, with GPT-4o (Multi-Gen) ranking second. However, GPT-4o (Multi-Gen) performs poorly in terms of avoiding patent profanity, averaging over 33 flagged instances per disclosure. In contrast, Patentformer achieves the best performance on this metric, with an average of just 0.02 instances, followed by both AutoSpec agents. Notably, all top-performing methods utilize models fine-tuned on patent disclosures, underscoring the importance of domain-specific training for accurately replicating the style and structure of patent language.

The AutoSpec agent that uses only template items exhibits the smallest difference in language diversity compared to the gold specifications, followed by the full-outline AutoSpec agent. All other baseline methods, with the exception of Patentformer, produce disclosures with greater language diversity than the original specifications. Patentformer performs the worst on this metric, showing a diversity difference more than twice as large as the next closest method.

Llama 3.3 and GPT-4o (Single-Gen) underperform across all metrics. Both attempt to generate the entire specification in a single generation, which likely contributes to their reduced performance. This contrasts with the other methods, all of which incorporate some form of task decomposition in the drafting process. These findings show the importance of breaking the drafting task into smaller sub-tasks to improve output quality.

\begin{figure}
    \centering
    \includegraphics[width=0.48\textwidth]{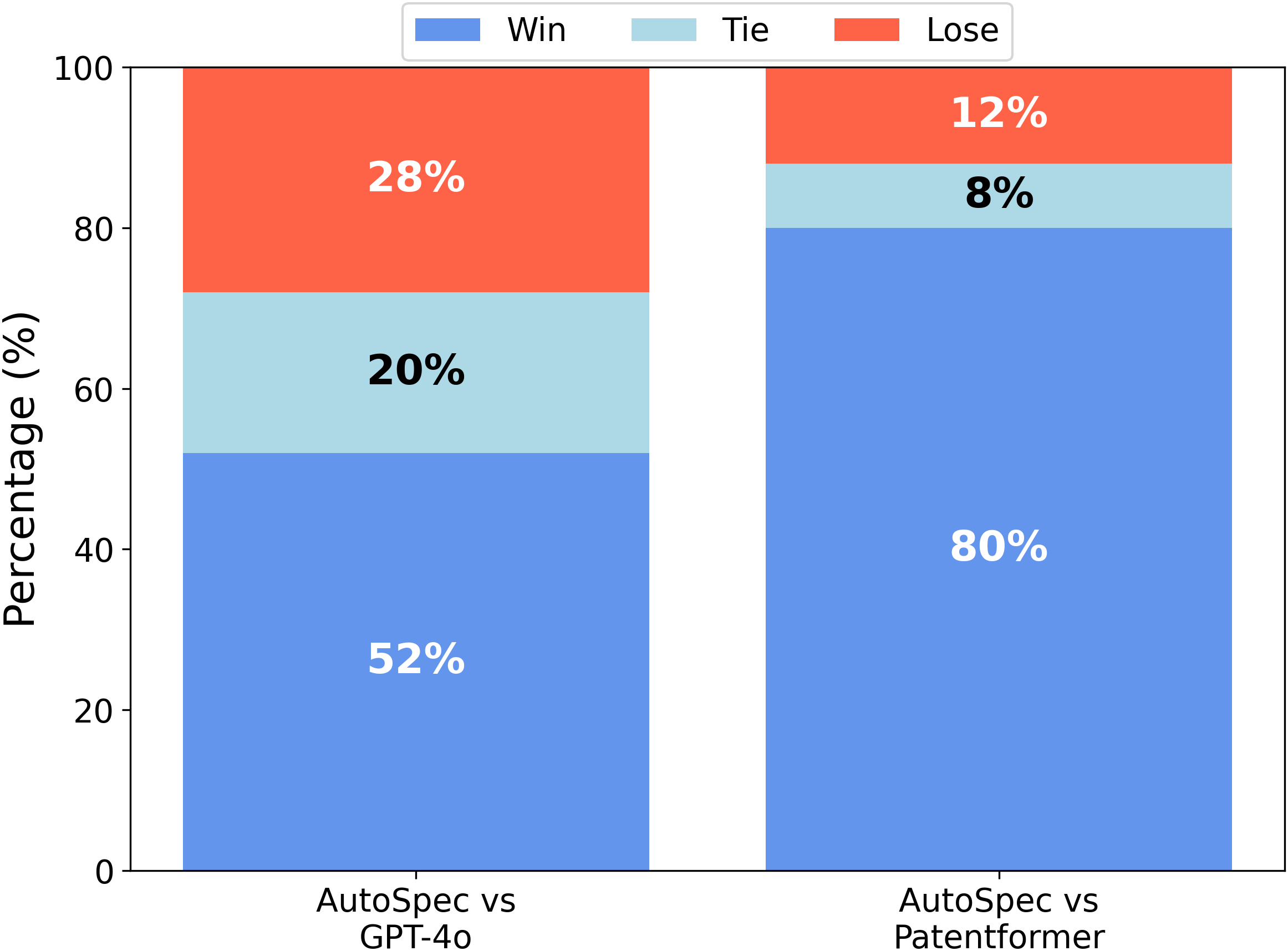}
    \caption{AutoSpec's win, loss, and tie rate vs GPT-4o and Patentformer according to expert rankings.}
    \label{fig:win_rate}
\end{figure}

\section{Expert Evaluation}

For our expert evaluation, we compared the performance of AutoSpec against Patentformer and GPT-4o (Multi-Gen). We generated patent disclosures for 25 biotechnology patents, selected by two experienced patent professionals. Each disclosure was evaluated using the annotation scheme described in Section \ref{sec:annotation_scheme}. In addition, the experts were asked to rank the disclosures based on their usefulness to a patent attorney as a first draft. To the best of our knowledge, this represents the first expert evaluation conducted on full, machine-generated patent specifications. We measured inter-annotator agreement using Kendall's Tau \cite{kendall_tau_1938} and obtained a score of 0.15, indicating a statistically significant correlation between expert ratings (see Appendix \ref{appendix:implementation} for details). 

The results of our expert evaluation are presented in Table \ref{tab:patent_eval}. AutoSpec outperforms all baselines across every metric except for elaboration, where GPT-4o achieves the highest score. Patentformer performs relatively poorly overall, though it demonstrates strong performance in language style, comparable to AutoSpec and notably better than GPT-4o. AutoSpec's respective win rates versus GPT-4o and Patentformer are given in Figure \ref{fig:win_rate}. Against GPT-4o, AutoSpec achieves a win rate of 52\% and a loss rate of 28\%. Its performance against Patentformer is even stronger, with a win rate of 80\% and a loss rate of only 12\%.

\subsection{Expert Comments and Error Analysis}

In addition to providing annotations and rankings, our expert evaluators also provided comments on each disclosure. They observed that the GPT-4o method frequently employed patent profanity, a finding consistent with our automatic evaluation. Specifically, the model often explicitly referenced claims and regularly used terms like “crucial” and “critical” to describe the invention. Experts also noted that its tone was overly conversational and tended to advocate for the invention rather than presenting it in the dry, factual style typical of patent specifications. This frequent use of advocating language sometimes led to incomplete explanations of claim elements, which affected it's ability to adequately address all of the claims in the disclosure.

Despite these shortcomings, GPT-4o demonstrated a notable strength in its ability to elaborate on claim concepts. It managed to do this effectively without relying on external tools such as internet search or retrieval mechanisms. This capability is likely attributable to the model's scale, both in terms of its size and the scope of its training data, which appears sufficient to support robust technical elaboration directly from its internal knowledge. This is in contrast to open-source models which have a more difficult time elaborating on technical concepts without leveraging external tools.

Experts commented that Patentformer had a tendency to hallucinate figures and certain aspects of the claims. It was often repetitive, frequently restating claim language without deeper elaboration. One area where Patentformer excelled was in it's language style, likely due to the model's extensive fine-tuning on patent disclosure. These results further highlight the importance of leveraging fine-tuning for automatic patent drafting. Patent specification is a unique instance where dry, technical language is highly desirable. This runs counter to the typical use cases for LLMs which are trained to be conversational and engaging. Therefore it is difficult for LLMs to emulate this language through prompting alone.

While the AutoSpec agent performed the best in general, there were notable failure modes highlighted during the evaluation. Many of these were centered around it's elaboration, which was the only evaluation category where it did not score the highest. More details on these shortcomings are in the Limitations section.

\section{Future Work}

One promising direction of future work is to extend AutoSpec to draft other sections of patent applications such as the claims.  Existing approaches to claim drafting typically generate claims either from prior claims or directly from the specification. However, this does not reflect the real-world drafting process, where patent attorneys often base claims on input provided by inventors. Incorporating this workflow into AutoSpec could lead to a more effective agent which can effectively generate both the claims and specification based on inventor-provided invention details.


Another potential direction is to develop more robust automatic metrics based on our evaluation protocol. In particular, assessing how closely LLM-generated annotations align with expert ratings could improve evaluation quality. This, in turn, could support the use of online training methods, such as reinforcement learning, to further refine patent drafting models.

\section{Conclusion}

Patent applications are key to protecting intellectual property and driving technological innovation. However, many smaller entities and individual inventors face obstacles to patenting their inventions due to the significant costs associated with drafting a patent application. To alleviate these issues we proposed AutoSpec, an agentic framework for automatically drafting patent specifications. AutoSpec's design is informed by expert input and mirrors the structured approach patent attorneys use to draft disclosures. To evaluate our framework we developed a novel, expert informed evaluation protocol for evaluating full patent disclosures. We evaluated our method using automatic and expert evaluations and found that our AutoSpec agent outperforms existing baselines on a patent drafting task. Additionally, we release a dataset of machine-generated patent disclosures annotated according to our evaluation protocol, providing a valuable resource for further research.

\section*{Limitations}

Despite its strengths, the AutoSpec system has limitations. In our expert evaluations, we observed instances where AutoSpec took certain technical concepts in the invention out of context. For example, in one instance a claim set made reference to ``scaffolding'' which in the context of chemistry refers to the core structure of a molecular compound or a class of compounds. However, the system mistakenly included a section in the disclosure discussing scaffolding in the context of construction. To protect sensitive claim content, we exclude claim text from the internet search component, but this occasionally leads to the retrieval and inclusion of irrelevant or misleading information in the disclosure.

Another limitation is that AutoSpec is currently built around text-only language models. Prior research has demonstrated that extracting OCR text from patent drawings can enable accurate figure descriptions \cite{wang-etal-2024-patentformer, shukla2025patentlmmlargemultimodalmodel}. However, integrating multimodal models that can process both text and images would likely enhance the quality of the generated specifications, making this a promising direction for future development.

Both our framework and evaluation protocol were developed with input from patent attorneys who practice in the United States. Since patent standards and disclosure requirements vary across jurisdictions, AutoSpec may not generalize well to other countries’ legal frameworks. Further work is needed to adapt and evaluate the system for use in patent offices outside the United States.

\section*{Acknowledgments}
We would like to thank Medler Ferro Woodhouse \& Mills PLLC for their financial support of this research and for dedicating attorney time to the review and evaluation of our method. Their support and expertise were instrumental in assessing the practical utility of the system.

\bibliography{custom}

\clearpage
\newpage

\appendix

\section{Additional Implementation Details}
\label{appendix:implementation}

In this section we give further details on our implementation of AutoSpec, baselines, and evaluations. All of our source code, including prompts is available. We also release the data we use to train AutoSpec as well as our expert-annotated evaluation data\footnote{\url{https://github.com/ryanshea10/AutoSpec.git}}.

\subsection{AutoSpec}

For AutoSpec, we train a LLaMA 3.3 70b parameter base model for one epoch using LoRA fine-tuning on four NVIDIA RTX A6000 GPUs. We use a learning rate of 5e-06 and an effective batch size of 8 on a dataset containing 1,354 patents, 750 come from the publicly available HUPD dataset \cite{hupd_neurips} and 574 were scraped from Google patents. We quantize the model to four bits using k-means quantization for use at inference time. We use the OpenAI web search tool for internet retrieval and retrieve the top one document that matches the search query.

\subsection{Patentformer}

We recreate the Patentformer method by training a LLaMA 3.3 70b parameter base model using the Patentformer dataset released by \citealp{wang-etal-2024-patentformer}. We use LoRA fine-tuning on four NVIDIA RTX A6000 GPUs for one epoch with a learning rate of 5e-05 and an effective batch size of 8. We quantize the model to 4 bits using k-means quantization for inference.

The Patentformer dataset consists of claims mapped to single paragraphs in the specification. Certain specification paragraphs also have OCR-extracted figure texts mapped to them. The dataset also includes the previous paragraph in the disclosure for context. The model is trained using this dataset to generate single specification paragraph from a single claim and the previously generated paragraph. The patentformer dataset includes tags to provide additional context to the model for generating specification, we remove these from the text during our final evaluations. See \citealp{wang-etal-2024-patentformer} for complete details.

\subsection{GPT-4o and LLaMA 3.3}

To create the GPT-4o and LLaMA 3.3 baselines we use prompt engineering to create the final bots. We focused on prompting the bots to adopt a legal language style and to format their disclosure without any markdown. GPT-4o was particularly prone to generating specification with markdown headers and lists. Whereas patent disclosure should be formatted as a series of paragraphs. Prompt engineering alleviates this to some degree, however GPT-4o still occasionally generates markdown in it's disclosure even explicitly told not to. This is also the case for language style. Both GPT-4o and LLaMA 3.3 struggle to replicate the language style of patent applications despite extensive prompt engineering and in-context examples.

\subsection{Semantic Similarity}

To measure semantic similarity we use two different SentenceTransformers \cite{reimers2019sbert_sentencetransformers} models that have been trained extensively on patent data. Both of these models have a maximum sequence length of 512 tokens, which is below the length of a typical patent disclosure. To measure the similarity between the full disclosures, we segment the document into smaller chunks, create embeddings for each chunk, then combine the embeddings using mean pooling \cite{abdaoui2023sbert_doc_embeddings}. Prior work has shown that specially trained models tend to be more effective at measuring sematic similarity than n-gram based metrics \cite{herbold2024semanticsimilaritypredictionbetter}. Therefore we choose this method for measuring semantic similarity as opposed to other metrics such as BLEU or ROUGE.

\subsection{Patent Profanity}

To measure patent profanity we look for the following terms within the disclosure: ``crucial'', ``critical'', ``prior art'', ``necessary aspect'', ``necessary component''. We also look for the term ``claim'' followed by an integer to assess where the model directly references the claims. These terms were provided to us by patent attorneys.

\subsection{N-Gram Diversity}
N-gram diversity is defined as the ratio of unique n-gram counts to all n-gram counts in a document \cite{shaib2025standardizingmeasurementtextdiversity}. We calculate the n-gram diversity for each disclosure using the following formula:
\[
\text{NGD}(D) = \sum_{n=1}^{10} \frac{\# \text{ unique } n\text{-grams in } D \oplus}{\# \text{ } n\text{-grams in } D \oplus}
\]

\subsection{Expert and Automatic Evaluations}

For both our expert an automatic evaluations we relied on patent attorneys to select our evaluation sets. This was done for the expert evaluation to ensure that the attorneys had the expertise to assess the patents. We also did this for the automatic evaluation to ensure a quality selection of patents. Not all published patents are of equal quality, and one key feature of a good patent application is its ability to withstand litigation. The attorneys we collaborated with have a strong track record of doing this therefore we chose to have them select our automatic evaluation set as opposed to randomly selecting patents from an existing dataset. Our data collection protocol is IRB approved.

During the expert evaluation we presented the attorneys with three different disclosures without telling them which specification was generated by which method. They rated each disclosure according to our evaluation protocol then ranked them based on how useful they would be if given to them as a first draft of a disclosure. These rankings are used to determine our win-rate in Table \ref{fig:win_rate}.

To measure the inter-annotator agreement between our raters we used Kendall’s Tau. Kendall's Tau is a measure of correlation between two sets of ordinal data, ranging from -1 to +1. A value of +1 indicates perfect agreement in rankings, -1 indicates perfect disagreement, and 0 indicates no association. This value can also be used for a statistical test with a null hypothesis of no correlation between the rankings.

Both patent attorneys annotated five overlapping patents during the evaluation which as used as the data for calculating inter-annotator agreement. We calculated Kendall’s Tau on our data and found a value of 0.15 which indicates a slight, but statistically significant correlation between the ratings for our sample size. We also measured the weighted Cohen's kappa for the rating and found a value of 0.17 which also indicates slight correlation \cite{cohen_kappa}.

\subsection{System Ablations}

We relied on small scale expert evaluations to test the different components of AutoSpec such as the retrieval tool, prompting methods, and inclusion of the different modules. This was done by generating two disclosures for each ablation and having one expert rate the outputs using our evaluation protocol from Section \ref{sec:annotation_scheme}. We used this method to determine the final design for our system. We found this method to be more reliable than automatic evaluations and also allowed us to elicit qualitative feedback from our experts.

The retrieval tool in particular is important for getting the model to elaborate on the technical information within the claims. Without it, model tends to simply repeat claim information without any elaboration. This can be seen in the reduced performance of the AutoSpec (Template) baseline which is designed as an ablation for the inclusion of the retrieval tool. Retrieval becomes less necessary when using more advanced models which are able to effectively elaborate on the claims without the need for external information. This is why no retrieval mechanism is included in the GPT-4o baseline.

Based on these evaluations, we found that the two most important components of our framework are the orchestrator and generator, which is expected given that these modules form the core of our method. The retrieval tool is the next most important given how important it is for encouraging elaboration. The merger tends to be the least impactful but is still important for ensuring the final description remains coherent. Ultimately, our framework consists of the minimal set of components required to generate high-quality patent specification as determined by these evaluations.

\end{document}